\documentclass[letterpaper, 10 pt, conference]{ieeeconf}  

\IEEEoverridecommandlockouts                              

\usepackage{graphicx}
\usepackage{amsmath}
\usepackage{mathtools}
\usepackage{todonotes}
\usepackage{wrapfig}
\usepackage{cite}
\title{\LARGE \bf
A Simple Approach for General Task-Oriented Picking \\ using Placing constraints}

\author{Jen-Wei Wang$^{*,1}$, Lingfeng Sun$^{*,1}$ Xinghao Zhu$^{1}$, Qiyang Qian$^{1}$ and Masayoshi Tomizuka$^{1}$
\thanks{* The authors contributed equally to this work. Author ordering determined by coin flip.}
\thanks{$^1$Department of Mechanical Engineering, 
        University of California, Berkeley, Berkeley, CA 94720, USA.
        {\tt\small {lingfengsun, jwwang, zhuxh, qiyang\_qian, tomizuka}@berkeley.edu}}
        }

\begin{document}

\maketitle
\thispagestyle{empty}
\pagestyle{empty}

\begin{abstract}

Pick-and-place is an important manipulation task in domestic or manufacturing applications. There exist many works focusing on grasp detection with high picking success rate but lacking consideration of downstream manipulation tasks (e.g., placing).
Although some research works proposed methods to incorporate task conditions into grasp selection, most of them are data-driven and are therefore hard to adapt to arbitrary operating environments.
Observing this challenge, we propose a general task-oriented pick-place framework that treats the target task and operating environment as placing constraints into grasping optimization. Combined with existing grasp detectors, our framework is able to generate feasible grasps for different downstream tasks and adapt to environmental changes without time-consuming re-training processes. Moreover, the framework can accept different definitions of placing constraints, so it is easy to integrate with other modules.
Experiments in the simulator and real-world on multiple pick-place tasks are conducted to evaluate the performance of our framework. The result shows that our framework achieves a high and robust task success rate on a wide variety of the pick-place tasks.

\end{abstract}

\section{Introduction}
Grasping is one of the most essential skills in robotic manipulation. It allows robots to perform various types of skills in different working environments starting from ``picking'' the target object. In recent years, we have seen tremendous progress in improving the grasp success rate of arbitrary new objects benefiting from large-scale datasets \cite{depierre2018jacquard}, randomization in simulation \cite{tobin2017domain,james2019sim}, and supervised deep learning methods \cite{bicchi2000robotic,redmon2015real,mahler2017dex,morrison2018closing, mousavian20196, breyer2021volumetric, zhu20216dcgpn, zhu2022learn}. However, there still remains a huge challenge of using grasping as a general tool in manipulation tasks where post-grasp tasks are involved. Despite the high success rates of existing methods for ``picking'' objects, many researchers \cite{fang2020learning,gupta2019relay,kalashnikov2018qt,lee2021ikea} still train the grasping skill from scratch when dealing with ``pick-and-place'' manipulation tasks. The reason is that a stable grasp, a grasp that can successfully pick up the target object without causing any slippage, does not guarantee the success of the follow-up tasks. In practice, stable but infeasible grasps might block the robot from finishing the following tasks. For instance, in Figure 1, we show that placing the same mug in different target environments requires the robot to grasp it with different poses. To learn a task-oriented grasping framework not only generalizable to objects but also to post-grasp tasks, we need to consider more than just the grasp quality.

The key challenge in task-oriented grasping is the object-task relationship and the accompanying task representation problem. Since not all the stable grasps for the object are feasible for the post-grasp tasks, we have to find feasible grasp poses for the corresponding tasks.
For pure grasping, the quality of grasp poses is mainly determined by the geometric properties of objects, and the generalization between objects comes from the shared geometric representation \cite{wang2022robot}. Deep networks learn the representation of visual inputs, benefit from domain randomization and large datasets, and provide stable quality predictions of poses. However, in task-oriented grasping, the feasible grasp poses of objects are determined by extra factors, including \emph{object affordance} \cite{detry2017task,kokic2017affordance,lakani2018exercising,do2018affordancenet} and \emph{operating environments}.

\begin{figure}[tb]
\begin{center}
	\includegraphics[width=0.48\textwidth]{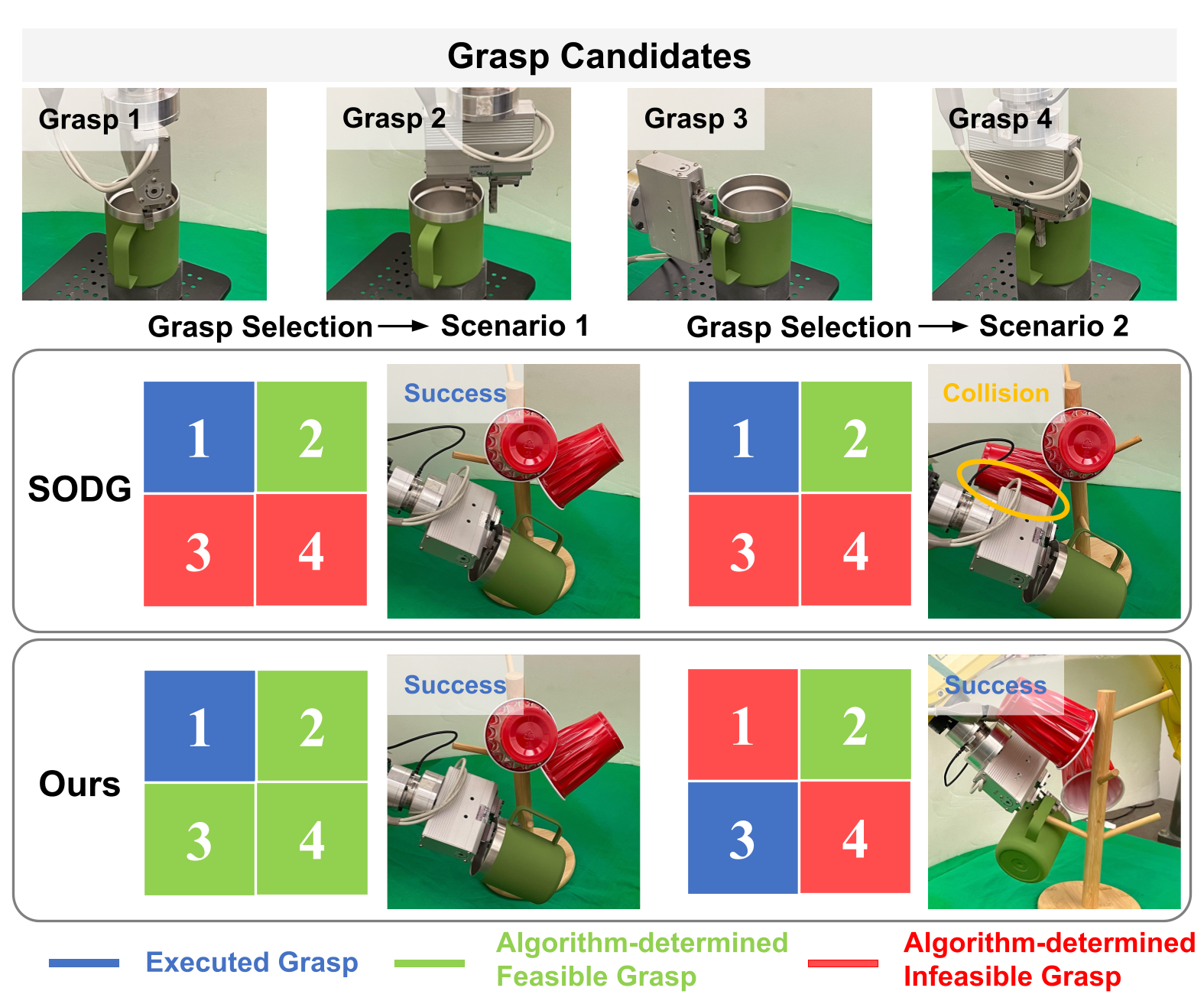}
	\caption{Compared to the previous approach \cite{murali2021same} (SODG), our proposed task-oriented grasp framework has a better generalization ability to different operating environments. After the scenario is changed (e.g., changing the hanging position of the red cup), SODG still labels the grasps on the handle part as infeasible and the grasps on the rim part as feasible, which leads to failure in scenario 2. On the other hand, our proposed algorithm can determine a feasible grasp based on the current scenario, resulting in task success in both scenario 1 and scenario 2.}
	\label{fig: comparison}
\end{center}
\vspace{-0.25in}
\end{figure}

Object affordance gives a prior distribution of feasible grasp poses based on the object name/type. When objects are designed for some specific functionalities, there are grasp poses related to those functions. CatGrasp \cite{wen2022catgrasp} proposed a framework to collect such prior distribution for a specific task and transform the distribution into other objects within a category. The actual use cases of objects further constrain the grasp poses, referred to as the ``same object, different grasps'' (SODG) setting \cite{murali2021same}, where multiple tasks are related to different tasks and feasible grasps are found conditioning on the task names.
Operating environments are more practical but essential constraints in real-world robotics tasks. Most manipulation tasks have objects in the environments to interact with, even for the same use-case. Because of the environmental change, the same object under the same task (use-case) may have different kinds of grasps. In Figure 1, the robot is operated on the same object (mug) in the same use case (hanging). However, different scenarios (operating environment) will lead to different feasible grasps.

\begin{figure*}[t!]
    \centering
    \includegraphics[width=0.9\textwidth]{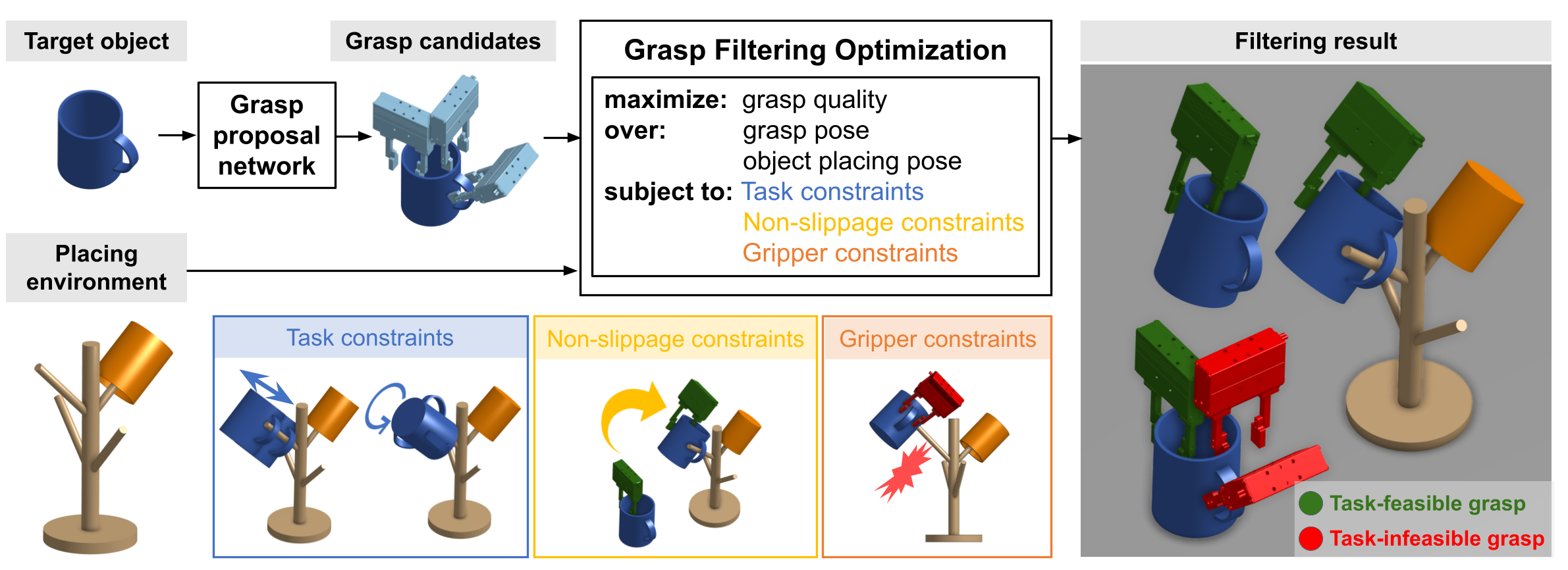}
    \caption{Our proposed framework of the task-oriented grasping. The grasp detector proposes diverse grasp candidates on the mug and our proposed optimization algorithm will determine the feasibility of each grasp candidate (red color represents infeasible grasp and green color represents feasible grasp) and execute the optimal grasp to hang the mug onto the rack. In the optimization algorithm, the objective is to select the grasp with the highest grasping quality and the constraints are to follow the placing constraints provided by user or learning-based model \cite{simeonov2022se}, non-slippage requirements of the object-gripper and the grasping feasibility between the gripper and environment}
    \label{fig:framework}
\vspace{-0.15in}
\end{figure*}

In practice, object affordance can be utilized in existing grasp methods using data-driven methods. The type and use cases of objects are usually finite and can be verbally described. Labeling the grasp with these task representations can provide task-oriented grasp proposals. However, environmental factors are harder to be defined as labels. In most cases, the feasibility of grasp poses depends on how the object is supposed to interact with the environment to achieve the goal. 
As a result, some researchers \cite{fang2020learning,gupta2019relay,kalashnikov2018qt,lee2021ikea} use imitation learning or reinforcement learning to learn the full policies of specific tasks instead of using an existing grasping pipeline as a starting point. 

How can we make a task-oriented grasp framework more general for post-grasp manipulation tasks? Although data-driven grasp proposal network, conditioned on the target task, works well in selecting task-oriented grasps by object and task labels, we find ``placing constraints'' of objects a good representation of task-related environments. Placing constraints refer to requirements on the object in order to finish the task, for instance, the final and key positions of the object in the environment during the placing operation. The picking grasp pose is related to the ``placing constraints'' by the non-slippage requirement between the robot pose and the object state. In this work, we propose a task-oriented grasp optimization framework using object placing constraints.


The contribution of this paper is summarized as follows:
\begin{itemize}
    \item A task-oriented grasping framework is proposed to determine task-achievable grasps and can generalize well to different pick-place tasks. 
    \item Our proposed framework can be adapted to task-related environmental change without additional data collection or re-training process.
    \item The framework is flexible to different kinds of placing constraints including the placing sets from human's annotation or from learning-based model \cite{simeonov2022se}.
    \item The simulated and real-world results show that our proposed framework performs well on the three testing tasks and the task success rate is bounded by how perfect the placing constraints will be.
\end{itemize}

\section{Related Works}
\subsection{Task-agnostic grasping}
Grasp detection approaches can be classified as two kinds: model-based approach and model-free approach. In model-based approach, CAD model of the target object is required and there are two kinds of methods that can generate grasp points on the object's geometry. One \cite{bicchi2000robotic} is to sample multiple grasp candidates on the object and use force-closure analysis as a quality measurement. The other one \cite{wang2022robot} is to offline collect grasp points on the CAD model and use 6D pose estimation to transfer collected grasps onto the target object. In model-free approach, grasp detectors can be grouped as 3-DoF grasp detectors and 6-DoF grasp detectors. For 3-DoF grasp detectors \cite{mahler2017dex, morrison2018closing,redmon2015real,zhu2022learn}, the model will predict top-down grasps on the target object based on RGB or depth image. However, it is not enough for top-down grasps to pick up some objects with complex geometries. Therefore, some research works propose 6-DoF grasp detectors \cite{mousavian20196, breyer2021volumetric,zhu20216dcgpn, jeng2021gdn} that can predict grasp points with different approaching directions to the target object. Although 6-DoF grasp detectors can already detect grasp points with diverse orientations, there are still few grasp points on partially-occluded region. Therefore, in this paper, we use multi-view planar grasp detector that can generate a diverse set of grasps covering the whole geometry of the target object. 

\subsection{Task-oriented grasping}
Task-oriented grasping approaches will decide grasp points that can both pick up the target object while achieving the task. Some research works \cite{detry2017task,kokic2017affordance,lakani2018exercising,do2018affordancenet} propose models to obtain the object affordance and use the affordance as a post-grasp task success rate. Nevertheless, such affordance learning approaches cannot generalize to different objects within a category. Therefore, CatGrasp \cite{wen2022catgrasp} proposed a category-level dense correspondence method for transferring affordance to different category-level objects. To generalize to unseen objects, SODG \cite{murali2021same} proposed a model that can decide the task success rate on each grasp candidate. Research works from computer vision community \cite{huang2015we,shan2020understanding,kokic2020learning} focus on pose estimation of human's hand when a human is trying to grasp an object and use that information to predict object affordance. On the other track of task-oriented research, previous work used self-supervised approaches \cite{fang2020learning} to jointly learn the task-oriented grasp detector and manipulation policy in a simulator. Another learning-based approach \cite{simeonov2022se} is to learn the policy from pick-place demonstration and use Neural Descriptor Fields (NDFs) \cite{simeonov2022neural} to prevent the model from losing generalization ability. However, none of these works focus on how to generalize to different operating environments. That is, if the target task or the operating environment is changed, they need to re-train the affordance map or manipulation policy, resulting in a time-consuming process. In this paper, we propose a flexible framework that can handle this issue and generalize to different tasks and environments.

\section{Problem Formulation}
\subsection{Notations}
\subsubsection*{Robot and object representation}
We use $g\in \textit{SE}(3)$ to denote a 6-DoF grasp pose for the gripper and $x\in \textit{SE}(3)$ to represent the position and orientation of a rigid body object. In a manipulation task, $x_0$ and $g_0$ are the initial state of the object and the picking grasp we want to solve. $x_f$ and $g_f$ are the corresponding final state of the object and gripper in the environment to finish the task. The regions occupied by $g_0$ and $g_f$ are denoted as $O_{g_0}$ and $O_{g_f}$ respectively. 
\subsubsection*{Task and environment representation}
The provided object placing set is denoted as $S_f$, which represents a set of possible object poses to finish the task. The operating environment is denoted as $e$ and the regions occupied by $e$ are denoted as $O_{e}$. Q represents the picking quality of a grasp (higher picking quality indicates a higher picking success rate).
\subsection{Placing constraints}
The placing constraints consist of two parts: \textit{task constraints} and \textit{gripper constraints}. Task constraints are the constraints for a target object to satisfy the completion condition of the task. For instance, in a mug-hanging task, a mug's handle should be placed on a stick of a rack. In a peg-in-hole task, a peg should be inserted into a hole. The constraints usually describe the state requirements for the objects. There are many ways to define task constraints, and we regard the task constraints as the possible pose sets $S_f$ for the target objects $x_f$. Gripper constraints are also important for placing since we adopt the gripper to manipulate the object to obey task constraints. In this paper, the gripper constraints are defined as the collision-free condition between the gripper and the operating environment.
\subsection{Task-oriented grasp proposal} \label{optimization_formulation}
The objective of solving our pick-and-place problem is to find a feasible high-quality ``picking'' grasp pose under the constraints created by the ``placing'' operation. We formulate the whole problem into a grasp filtering optimization:
\begin{subequations}
\begin{align}
\max_{g_0, x_f} \quad &  Q(g_0, x_0) \label{mainopt:obj} \\
\textrm{s.t.}  \quad 
& g_f = (x_f*x_{0}^{-1})*g_0 \label{mainopt:c4} \\
& \phi(O_{g_0}, O_{e})\geq d_{safe}\label{mainopt:c1}\\
& \phi(O_{g_f}, O_{e})\geq d_{safe}\label{mainopt:c2}\\
& x_f \in \mathcal{S}_f \label{mainopt:c5}
\end{align}
\end{subequations}
where the optimization variables are the gripper's picking pose $g_0$ and the object's placing pose $x_f$. (\ref{mainopt:obj}) is the target grasp quality to maximize, with $x_0$ and $x_f$ representing the initial picking state and the final placing state of the object. (\ref{mainopt:c4}) requires the placing grasp pose is the estimated placing grasp pose calculated using $g_0, x_0, x_f$ due to non-slippage requirements of the object and the gripper. Each grasp pose or object pose is the 4x4 homogeneous transformation matrix in world coordination. $x_f*x_{0}^{-1}$  will calculate the rigid-body transformation from initial object pose to final object pose and this transformation will be applied to $g_0$ to obtain $g_f$. (\ref{mainopt:c1}) and (\ref{mainopt:c2}) describe the collision constraints at initial and final grasp pose. (\ref{mainopt:c5}) represents the task constraint that the final object state $x_f$ should lie in the object placing set $S_f$.


\section{Methods}
\subsection{Overall pipeline}

\begin{wrapfigure}{R}{0.2\textwidth}
\centering
\includegraphics[width=0.2\textwidth]{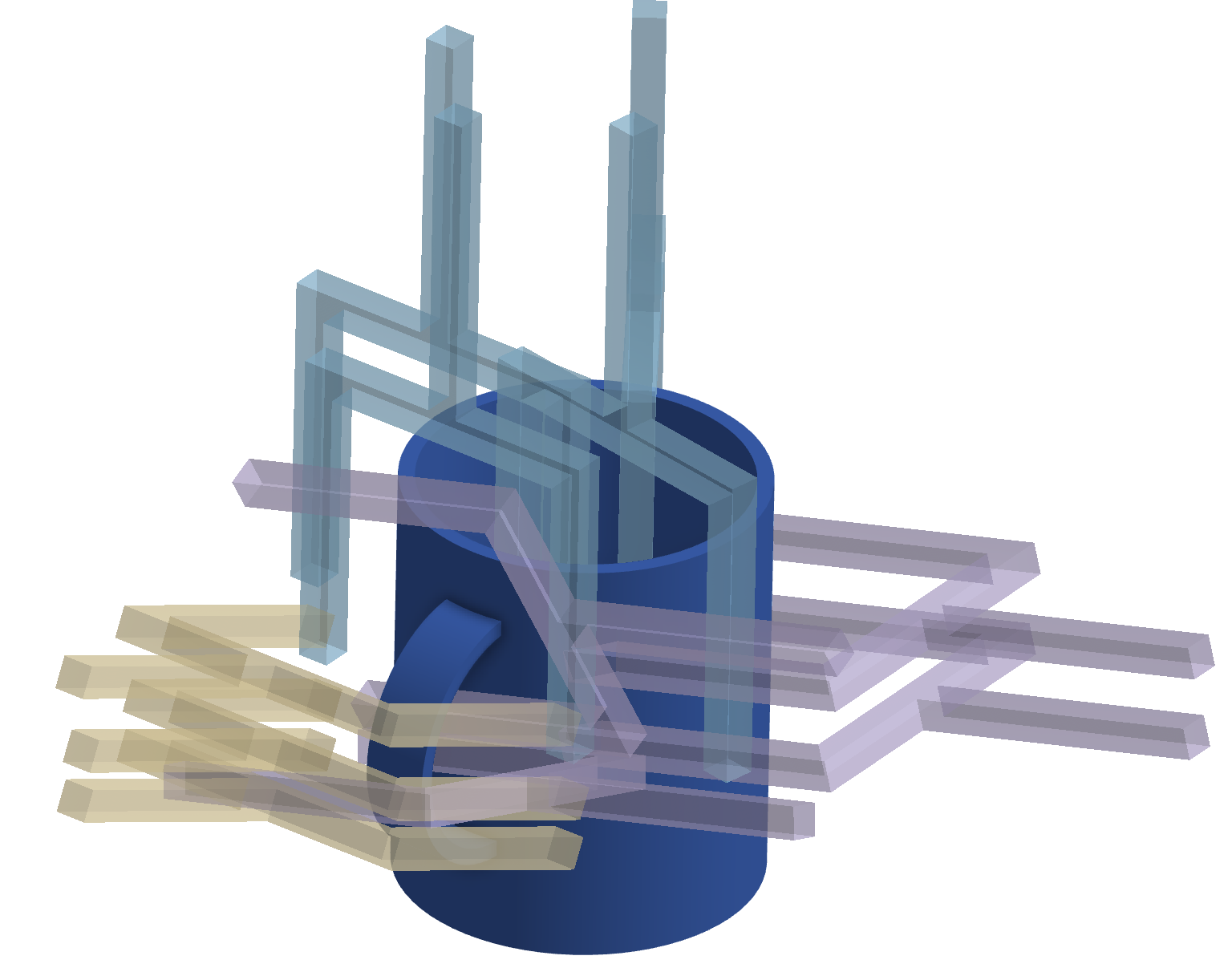}
\caption{\label{fig:diverse}Diverse grasp candidate generated from multi-view grasp detector. Different color represents the detected grasps from different view image.}
\end{wrapfigure}

To solve the optimization problem defined in Section \ref{optimization_formulation}, we propose a sampling-based pipeline to find the optimal grasp point $g_0^*$ and final object state $x_f^*$ given a provided placing set $S_f$ and operating environment $e$. As shown in Figure \ref{fig:framework}, the overall pipeline mainly consists of two modules, grasp proposal network and grasp filtering optimization. In grasp proposal network, a multi-view grasp detector is adopted to provide diverse sets of grasp candidates on a target object as shown in Figure \ref{fig:diverse}. However, not every grasp candidate is feasible for the target task. Therefore, grasp filtering optimization is adopted to filter out the infeasible grasps and choose the optimal grasp based on the target task and operating environment. 

To filter out the infeasible grasps, the optimization algorithm will first check the gripper constraint defined in Equation (\ref{mainopt:c1}) to make sure each grasp candidate is collision-free with the environment. Because task constraints are provided in Figure \ref{fig: set}, each picking gripper pose, obeying Equation (\ref{mainopt:c1}), can be mapped to placing gripper pose via the non-slippage constraint defined in Equation (\ref{mainopt:c4}). Once each picking gripper pose is mapped to placing gripper poses, we can check the gripper constraint defined in Equation (\ref{mainopt:c2}) to eliminate grasps that will cause collision during the placing process. Figure \ref{fig: rigid} visualizes the mapping and collision-checking process. In sum, with the help of the non-slippage property of the object and the gripper (Equation (\ref{mainopt:c4})), the task constraints (Equation (\ref{mainopt:c5})) can be transformed into gripper constraints (Equation (\ref{mainopt:c1}) and (\ref{mainopt:c2})). This property will help the algorithm adapt to different tasks and environments.

\subsection{Grasp Proposal Networks}

The grasp proposal network should provide diverse grasp candidates that can cover the whole geometry of a target object as shown in Figure \ref{fig:diverse}. The diversity of the grasp candidates is important for downstream optimization because diversity will increase the chance of finding task-achievable grasps. Although single-view 6-DOF grasp detectors can detect grasp points with different orientations, partial observation still limits the diversity of detected grasps. To detect diverse sets of grasp points for the downstream algorithm, we apply a planar grasp detector on three different views of a target object. After the detection, all planar grasp points on each view are mapped from their corresponding image coordinate to the robot's coordinate for aggregating all planar grasps together as shown in Figure \ref{fig:diverse}. In this work, the focus is not on the increase in the success rate or diversity of the detected grasps. Therefore, as long as the grasp detector can provide diverse grasp points with high grasping quality, it can be adopted to be the grasp proposal network.

\subsection{Grasp Filtering Optimization}

We adopt a sampling-based algorithm to realize grasp filtering optimization. The grasp proposal network has already determined the grasp quality of each candidate. The constraints defined in optimization formulation are used to examine whether each grasp candidate can be executed in the operating environment to achieve the target task. We classify the constraints into three kinds: task constraints, non-slippage constraints, and gripper constraints. The non-slippage constraint is the key to this optimization problem because it can transform the provided task constraints into gripper constraints. Specifically, we can map each grasp candidate to their corresponding placing gripper poses, so that we know whether each candidate grasp can achieve the task or not before actually executing that grasp.

\begin{figure}[tb]
\begin{center}
	\includegraphics[width=0.45\textwidth]{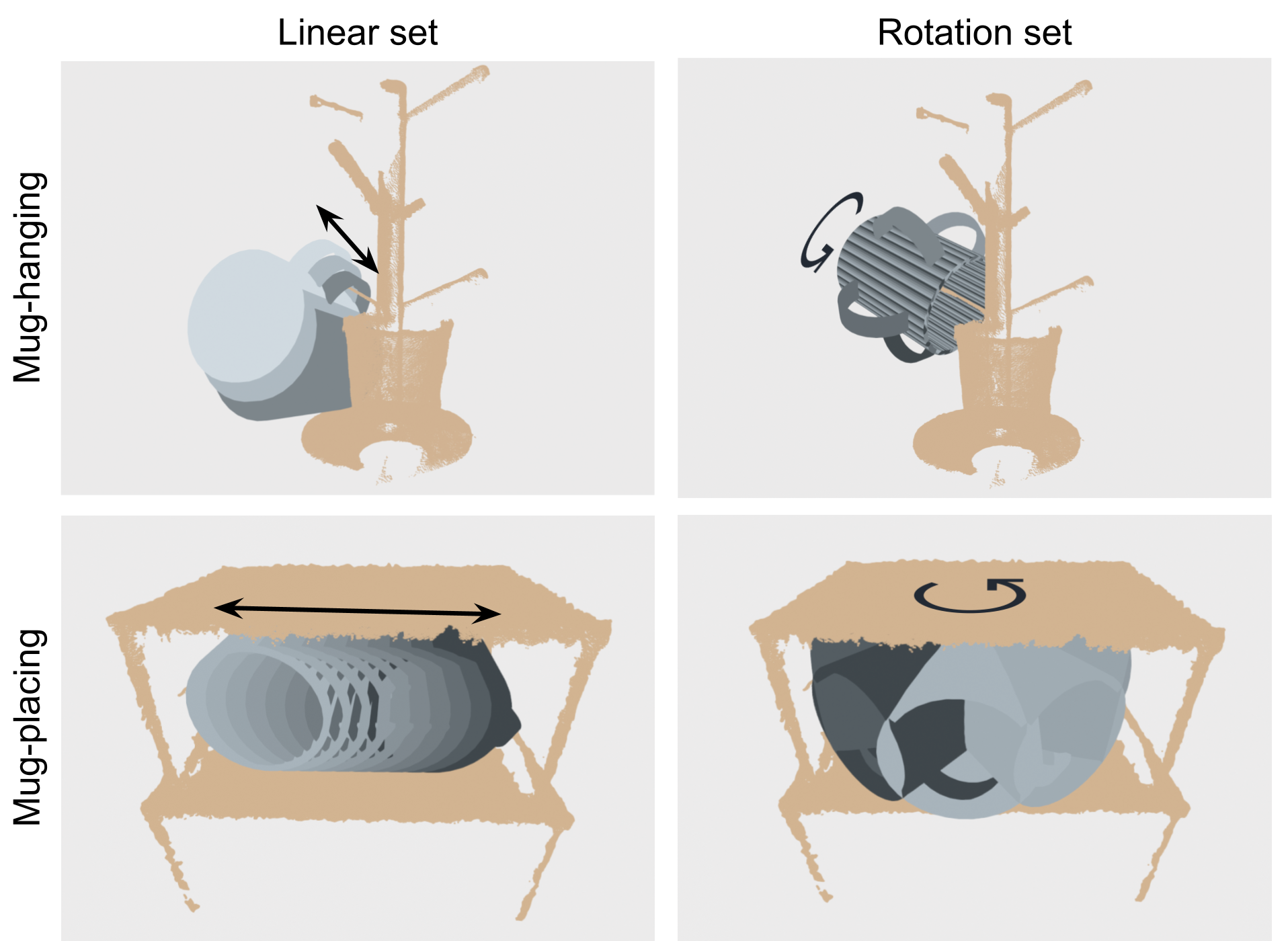}
	\caption{The task constraints (Equation (\ref{mainopt:c5})) in mug-hanging task and mug-placing task. The first row shows two kinds of placing sets in mug-hanging tasks. The linear placing set defines the placing poses where the handle of the mug should be hung on the stick, and the mug can be translated along the stick. The rotational placing set defines the placing poses where the mug can be placed upside down on the stick and can be rotated along the stick. The second row shows two kinds of placing sets in mug-placing task. The linear placing set defines the placing poses where the lying mug can be translated inside the shelf, and the rotational placing set is a set of placing poses where the lying mug can be rotated inside the shelf.}
	\label{fig: set}
\end{center}
\vspace{-0.15in}
\end{figure}

\begin{figure}[tb]
\begin{center}
	\includegraphics[width=0.45\textwidth]{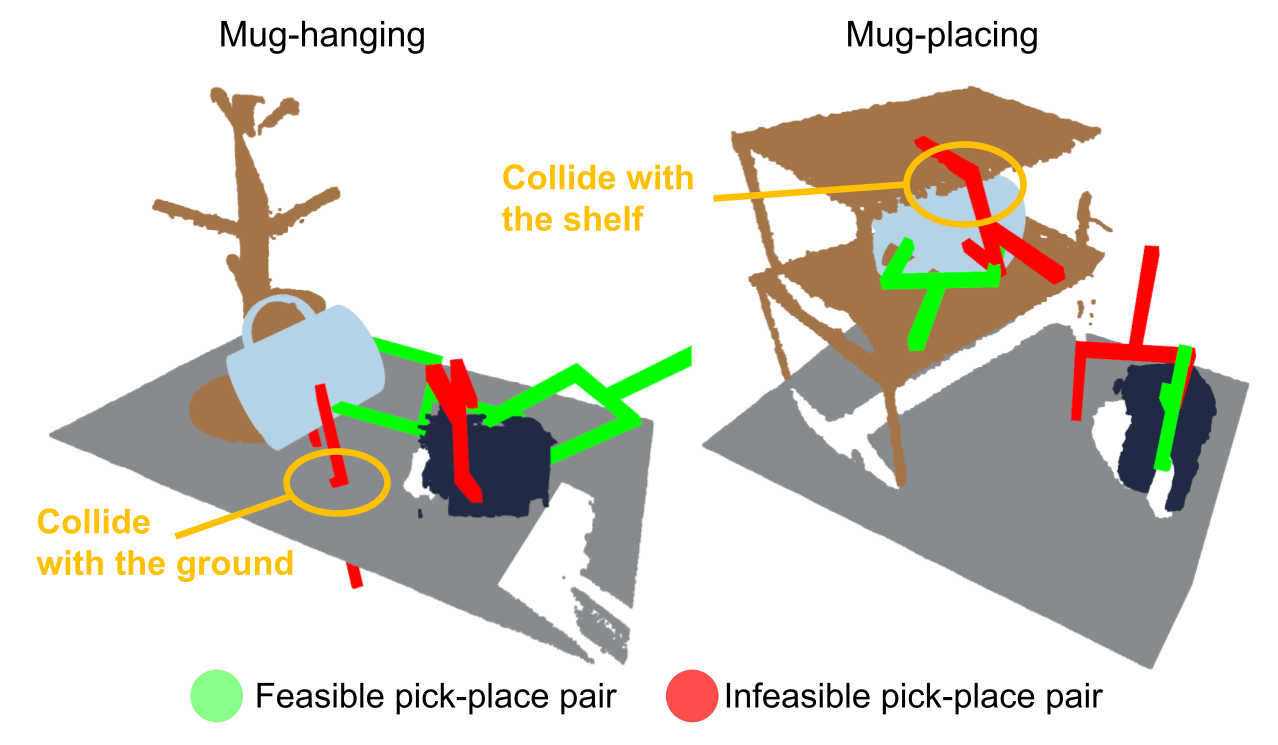}
	\caption{The non-slippage constraints (Equation (\ref{mainopt:c4})) and gripper constraints (Equation (\ref{mainopt:c1}) and (\ref{mainopt:c2})) in mug-hanging and mug-placing task. For each task, two viewpoints are provided in order to show two pairs of pick-place gripper poses. The optimization algorithm will first map each grasp candidate to a placing pose and examine whether each obeys the gripper constraint. The green pair is the task-feasible grasps, and the red pair is the task-infeasible grasps.}
	\label{fig: rigid}
\end{center}
\vspace{-0.25in}
\end{figure}

\subsubsection{Task constraints} \label{sub:task_const}

There are many ways to include task constraints in the optimization algorithm. In this paper, we depict task constraints as a possible set of object's final states, which is denoted as $S_f$ in Equation (\ref{mainopt:c5}). This set is defined by a human user or learning-based model \cite{simeonov2022se}. Figure \ref{fig: set} shows two examples of the task constraints. The linear placing set in the mug-hanging task is the set where the mug's handle part should be hung onto a stick, and the set is defined as the translation along the stick. The rotational placing set in the mug-hanging task is the set where the mug can be placed upside down onto the stick and the set is defined as the rotation along the stick. For the mug-placing task, the linear set is a set where the lying mug can be translated inside the shelf and the rotational set is a set where the lying mug can be rotated inside the shelf. Other than these manually defined sets, we also learn a model to automatically determine placing sets based on a few demonstration data.

\subsubsection{Non-slippage constraints} \label{sub:rigid_const}

We assume that there is no slipping between the object and the gripper during the pick-and-place process, so the object and the gripper will follow a similar rigid body transformation defined in Equation (\ref{mainopt:c4}). This constraint relies on the quality of the grasp detector. With the help of this relationship, we can take the task constraints mentioned in Section \ref{sub:task_const} into account. Specifically, each grasp candidate can be mapped to the corresponding placing state $g_f$ by applying the transformation between the object's initial state $x_0$ and its final state $x_f$ (shown in Figure \ref{fig: rigid}). Note that $x_f$ is not a single state. Instead, it belongs to a placing set $S_f$. Therefore, each picking gripper pose can be mapped to several placing gripper poses. Once one of the placing grippers poses obeys the gripper constraints in Equation (\ref{mainopt:c2}), the corresponding picking gripper pose will be regarded as task-achievable. In sum, the definition of placing set relaxes the condition of a grasp to be considered task achievable.  

\subsubsection{Gripper constraints}

Depending on the target task, gripper constraints can be defined in many ways. For general placing tasks, a common way to determine the gripper constraints is to check the collision between the gripper and the environment. In this project, we mainly target on placing tasks, so we use the Signed Distance Field (SDF) \cite{oleynikova2016signed} of the environment to check the feasibility of a grasp.
In practice, we need to have clearance between the gripper and the environment. Therefore, the gripper constraint is defined as Equation (\ref{mainopt:c1}) and (\ref{mainopt:c2}), where $d_{safe}$ denotes the clearance between the gripper and the environment.

\section{Experimental setup}

In this section, we test the effectiveness of our proposed pipeline in multiple manipulation task settings using both simulations and real-world experiments. We define three standard pick-place manipulation tasks where the success of the task is sensitive to the placing pose of the objects.
To complete the task, both grasping and post-grasp planning are required.
To evaluate the generalization ability of our proposed framework, each task consists of five cases where the placing set $S_f$, the initial state of a target object $x_0$, and the operating environment $e$ differ. Each execution is regarded as a success if the robot can pick up an object and achieve the task both in simulated and real-world experiments.

Using these experiments, we would like to demonstrate that our proposed framework can 
\textbf{\emph{i)}} determine task-oriented feasible grasp poses in general pick-place tasks,
\textbf{\emph{ii)}} zero-shot adapt to scenario differences and environment changes
and \textbf{\emph{iii)}} work for different types of placing constraints.

\begin{figure}[tb]
\begin{center}
	\includegraphics[width=0.45\textwidth]{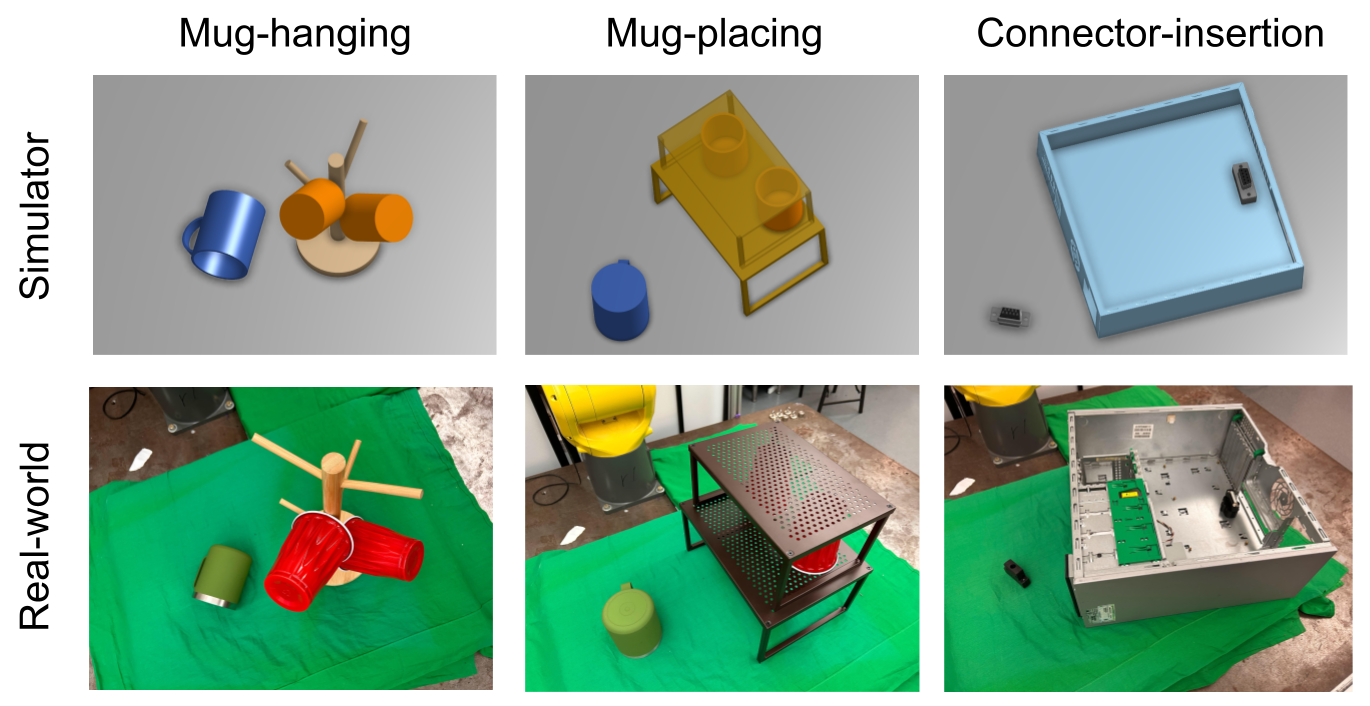}
	\caption{We evaluate three kinds of testing tasks in simulator and real-world. For each task, the initial state of the target object and the operating environment are changed in order to create different scenarios. In mug-hanging task, a robot is asked to hang the mug onto a rack. In mug-placing task, the mug is about to be placed onto a shelf. In connector-insertion task, a robot is asked to insert the connector into the socket.}
	\label{fig: exp}
\end{center}
\vspace{-0.25in}

\end{figure}

\subsection{Testing task} \label{sub:testing_task}

Figure \ref{fig: exp} describes three different testing tasks in the simulator and real-world. Task one is the \textit{mug-hanging} task; task two is the \textit{mug-placing} task, and task three is the \textit{connector-insertion} task.
Each task consists of multiple scenarios: the target object will be placed on the table in different initial poses, and the operating environment will be changed.
\begin{itemize}
    \item \textbf{mug-hanging}: pick up a mug (target object) and hang the mug onto different positions of a rack. The operating environment consists of a rack with some hanging cups. Environmental changes include the pose change of hanged cups.
    \item \textbf{mug-placing}: pick up a mug (target object) and place the mug in different positions on a shelf. Environmental changes include the position change of existing objects on the shelf in each scenario.
    \item \textbf{connector-insertion}: pick up a connector (target object) and insert it into a target socket inside a computer case. Environmental changes include design changes of sockets layout inside the computer case.
\end{itemize}

\subsection{System design}

We perform the evaluation in both simulation and real-world with similar settings. As shown in Figure \ref{fig: comparison} and Figure \ref{fig: exp}, the system consists of a Fanuc LR mate 200iD robotic arm with a two-jaw parallel gripper and two Ensenso 3D cameras. 
A target object and the corresponding operating environment are within the robotic arm's reachable range. For each scenario, a target object is randomly placed in front of the robotic arm, and the operating environment is changed according to the testing task mentioned in Section \ref{sub:testing_task}. Two 3D cameras are used to get the point cloud of the environment from different viewpoints. Once the point cloud is obtained, our proposed framework is used to predict multiple grasp candidates on the target object and choose the optimal grasp point for pick-and-place execution.

In our optimization framework, we need to estimate $x_0$ and $S_f$ so that the rigid body transformation can be obtained (described in Section \ref{sub:rigid_const}). We implement the pose estimation via ArTag detection. Once each scene is set up, we use ArTag to obtain the object's initial pose $x_0$. $S_f$ will be manually defined by ArTag or automatically defined by NDFs \cite{simeonov2022se}. Pose estimation is essential but not the main focus of our paper, one can use arbitrary pose estimation algorithms to obtain  $x_0$ and $S_f$ in different scenes.
\subsection{Evaluation metrics}
In each testing task, five different scenarios (two scenarios in simulator and three scenarios in real-world) are randomly sampled. In each scenario, there will be multiple grasp candidates on the target object. Therefore, for each task, there will be a number of initial grasp candidates. The task success rate and task recall metrics introduced below are based on these grasp candidates for each testing task. 

\begin{figure*}[t!]
    \centering
    \includegraphics[width=0.9\textwidth]{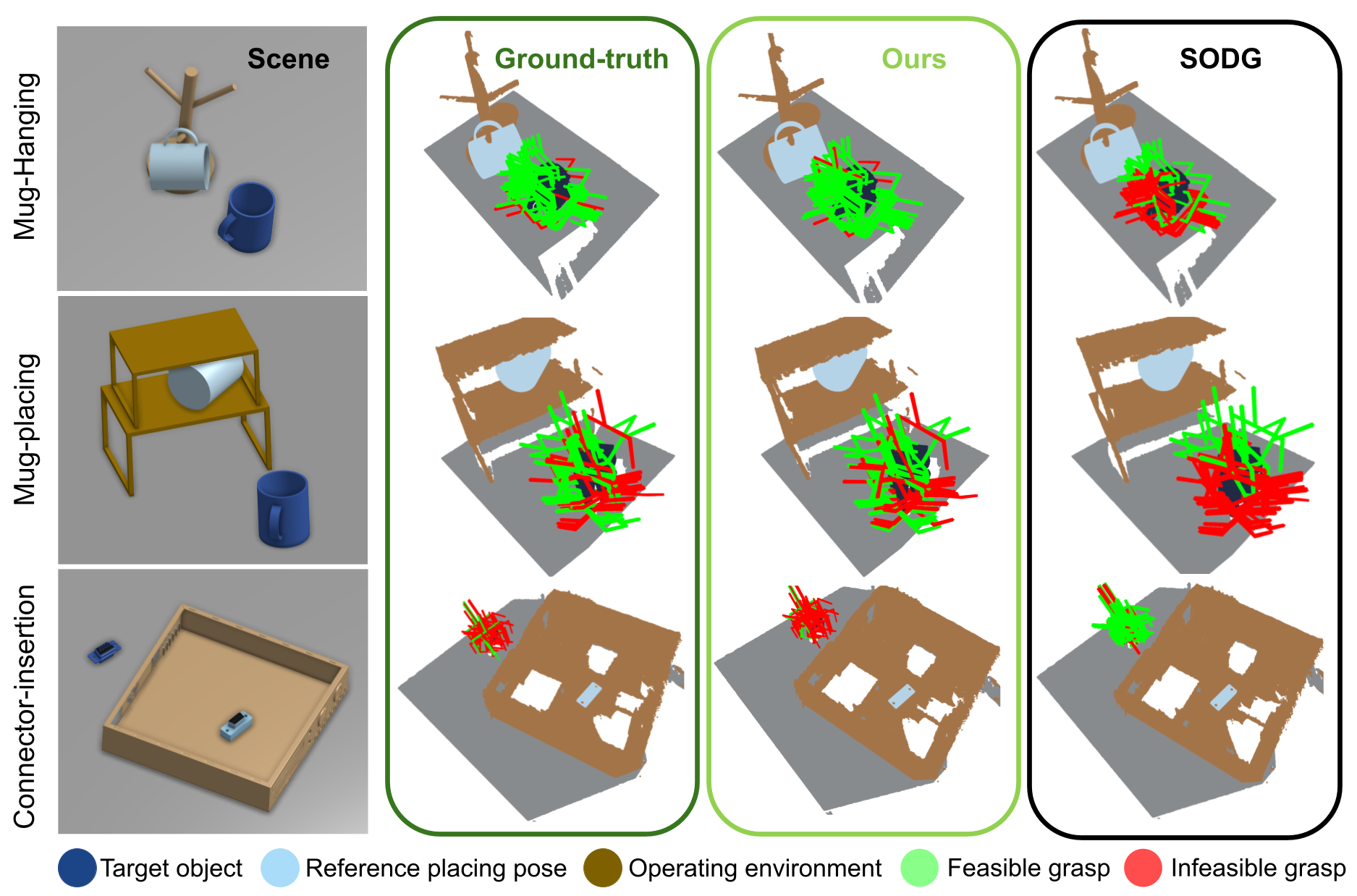}
    \caption{Grasp filtering result on three testing tasks. Based on point cloud data, the grasp proposal network will first generate diverse grasp candidates on the target object in each scene. Then, our proposed framework will determine whether each grasp candidate is \textit{task-achievable}. The green color indicates that the grasp point can help achieve the task, and the red color indicates that the grasp point will lead to task failure. Compared to the SODG column, the color distribution of our column is closer to the ground-truth column, meaning that our proposed algorithm can better determine whether each grasp candidate is feasible for the target task.}
    \label{fig:result}
\vspace{-0.1in}
\end{figure*}

We define two metrics: \textit{\textbf{task success rate}} and \textit{\textbf{task recall}} to evaluate the effectiveness of the framework. 
To calculate the metrics, we first need to acquire the ground-truth labels of all grasp candidates. To obtain the label shown in Figure \ref{fig:result}, showing if the grasp pose is feasible to complete the pose-grasp placing task, the robotic arm is asked to execute each grasp candidate to pick up the target object and try to achieve the target task(if it is successfully picked up).
As a result, each grasp candidate can be classified as \textit{picking-success} or \textit{picking-failure}. For each \textit{picking-success} grasp, it can further be classified as \textit{task-achievable} or \textit{task-failure}.

In general, the GG-CNN\cite{morrison2018closing} can provide picking grasps of around 83.77\% success rate in real-world experiments; this can be fine-tuned and improved with more advanced grasp proposal networks\cite{breyer2021volumetric,zhu20216dcgpn}. Since grasp detection is not the main focus of this paper, we calculate the task success rate and recall without the consideration of picking-failure grasp. Specifically, success rate and recall are defined as
\begin{equation} \label{equ:success_rate}
\textrm{Task\,success\,rate} = \dfrac{|executed\,set \cap achievable\,set|}{|executed\,set|}
\end{equation}
\begin{equation}
\textrm{Task\,recall} = \dfrac{|executed\,set \cap achievable\,set|}{|achievable\,set|}
\end{equation}
where the executed set is the set of grasps which are determined as executable by an algorithm (successfully picked). Note that the executed set already excludes the picking-failure grasp. The achievable set is the set of grasps which are labeled as \textit{task-achievable}.

In practice, the \textit{task success rate} is more important than \textit{task recall} since it greatly determines whether the algorithm can finish the task. However, if we would like to combine our framework with other downstream algorithms, such as force-based insertion, the algorithm with high task recall will provide more grasp choices for the downstream algorithm.

\subsection{Baselines}
\subsubsection{Vanilla} 
Basic pick-place framework using a grasp proposal network (GG-CNN) and does not take the downstream task into account. We include this vanilla method in the comparison to see the improvement in success rate by taking the influence of downstream tasks into consideration.
\subsubsection{SODG}
SODG\cite{murali2021same} determines the task-oriented grasp based on the trained grasp classify network on the target task. To train the network for task-oriented grasp, humans need to label each grasp candidate as task-feasible or task-infeasible based on their prior knowledge of the target task. For example, grasping the handle part of a mug has a higher failure chance (collision exists) during the hanging process in the mug-hanging task. Therefore, the algorithm will classify the grasps on the handle part as task-infeasible and the other grasp points as task-feasible. For the mug-placing task, the mug might need to lie down to be placed into the shelf. Therefore, the grasp points on the body of the mug are task-infeasible, and the other grasps are task-feasible. For the connector-insertion task, the top-down grasps will be considered as task-feasible due to the final insertion direction, and the other will be task-infeasible. After labeling the task knowledge on each grasp, we can learn the SODG network and predict task-oriented grasps. Compared with this object-centric approach, we would like to show that prior knowledge of the object-task relationship is helpful but might not be enough, especially in changeable environments, and our proposed framework can be flexible in avoiding placing constraints violations.

\section{Results and Discussions}
\subsection{Task Success Rate and Recall}
The experimental results of the simulated and real-world robot working on three tasks are shown in Table \ref{tab:precision} and Figure \ref{fig:result}. For each task, we evaluate three approaches on two scenarios in simulator and three scenarios in real-world. In each scenario, each approach will determine 10-12 executed grasps from the candidate grasps generated by the grasp proposal network. Therefore, the total number of executed grasps is 60 in mug-hanging and mug-placing task, and 58 in connector-insertion task. The vanilla method only considering the picking condition has very different performance in the three tasks, showing different sensitivity of picking pose in different tasks. 
In all three tasks, our proposed framework has a significant improvement over the baselines and reaches a high success rate.
The most common reasons for failure cases are the violation of non-slippage constraints and the sim-to-real errors in execution.
Based on Figure \ref{fig:result}, we can find that the mug-placing task is the most challenging one because the shelf will introduce hard constraints to the gripper and limit the options for gripper's orientation. Even the task success rate of vanilla method in mug-placing task is higher than the one in connector-insertion task, the pick-place transformation of the target object and the placing environment still complicate the grasp selection process. For the mug-hanging task, there are many grasps that are task-achievable, even for the vanilla framework. In this task, the SODG approach achieves an even worse success rate compared to the vanilla version since the placing constraints in this scene do not comply with human prior knowledge; some grasp points on the handle are success grasps. 
In terms of recall, we further evaluate whether we can find out all the task-achievable grasps defined in the numerator of vanilla method in Table \ref{tab:precision}. Based on Table \ref{tab:recall}, our proposed algorithm also has a better performance in terms of finding as many task-achievable grasps as possible. We can conclude that SODG is a more conservative algorithm in this case.
\begin{table}[h]
\begin{center}
\caption{Task Success Rate Results} \label{tab:precision}
\begin{tabular}{cccc}
\hline
                    & Vanilla & SODG \cite{murali2021same}  & Ours           \\ \hline
Mug-hanging         & 52/60   & 47/60 & \textbf{58/60} \\ \hline
Mug-placing         & 40/60   & 43/60 & \textbf{59/60} \\ \hline
Connector-insertion & 16/58   & 16/58 & \textbf{57/58} \\ \hline
\end{tabular}
\end{center}
\vspace{-0.1in}
\end{table}

\begin{table}[h]
\begin{center}
\caption{Task Recall Results} \label{tab:recall}
\begin{tabular}{cccc}
\hline
          & SODG \cite{murali2021same}  & Ours            \\ \hline
Mug-hanging      & 22/52 & \textbf{52/52} \\ \hline
Mug-placing     & 14/40 & \textbf{39/40} \\ \hline
Connector-insertion & 15/16 & \textbf{16/16} \\ \hline
\end{tabular}
\end{center}
\vspace{-0.15in}
\end{table}

In summary, we believe that semantic knowledge of the target object is just one way to define task constraints but has its limitations in real-world environment variations when prior knowledge of the object-task relationship does not match the real situation. In addition to the same-object-different-grasp assumptions, task-oriented grasp can benefit more from environmental task constraints. The task placing constraints, non-slippage constraints, together with gripper constraints integrate the operating environment information into our grasp filtering framework and can be generalized to different scenarios in each task with provided placing constraints.

\begin{table}[]
\begin{center}
\caption{Ablations on placing sets in terms of task success rate} \label{tab:ablation}
\begin{tabular}{ccccc}
\hline
        & \begin{tabular}[c]{@{}c@{}}Linear set\\ (L)\end{tabular} & \begin{tabular}[c]{@{}c@{}}Rotation set \\ (R)\end{tabular} & \begin{tabular}[c]{@{}c@{}}Ours\\ (L x R)\end{tabular} & \begin{tabular}[c]{@{}c@{}}NDFs\\ \cite{simeonov2022se}\end{tabular}  \\ \hline
Mug-hanging & 57/60                                                      & 58/60                                                       & \textbf{58/60}                                                      & 53/60 \\ \hline
Mug-placing & 14/60                                                      & 18/60                                                       & \textbf{59/60}                                                      & 57/60 \\ \hline
\end{tabular}
\end{center}
\vspace{-0.25in}
\end{table}

\subsection{Ablations on Placing constraints}

One important reason for our proposed approach to reach near-perfect performance in all tasks is the choice of placing sets as constraints for each testing task. To demonstrate the importance of constraint formats, we investigate on how different placing sets will influence the performance of our proposed algorithm. Based on Figure \ref{fig: set}, we define a linear set and rotation set for both mug-hanging and mug-placing task. Moreover, we add another set that is the Cartesian product of linear set and rotation set in the ablation study. We also evaluate the performance of the framework when the placing set is determined by learning-based model based on few demonstration data. Our selected approach \cite{simeonov2022se} will determine local coordinate on task-related objects (e.g. mug and rack in the mug-hanging task) via Neural Descriptor Field (NDFs) \cite{simeonov2022neural} and decide the pick-place transformation by learning how two objects align together from expert's demonstration. To implement this approach, a model is firstly trained on 5 demonstrations for each task, and predict the placing set on each testing scenario. As a result, the predicted placing poses are close but not perfectly aligned to the placing poses in demonstrations. Note that there is only one possible placing pose in the connector-insertion task, so we do not do ablation study on this task.

From Table \ref{tab:ablation}, we observe that defining more flexible constraint set for challenging tasks such as \textit{mug-placing} is the key to the high success rate. If we only define a linear set or rotation set in the mug-placing task, most of the placing grasp poses will collide with the shelf, resulting in low success rate. In NDFs, we cannot demonstrate for all possible placing poses, so the success rate will be bounded by the diversity of the demonstration.
In summary, the overall success rate is sensitive to the definition of placing constraints. A proper definition of target placing pose set contributes a lot to the performance. 
Learning more general placing constraints using O2O-Afford \cite{mo2022o2o} is one of our future directions.



\section{Conclusion}

We propose a general task-oriented grasping framework that can incorporate task conditions into placing constraints and provide task-feasible grasps that help robot achieve pick-place tasks. This framework can handle environmental change and find optimal grasps under different scenarios. Moreover, our proposed framework works well on different kinds of placing constraints. In the simulated and real-world experiments, our framework outperforms SODG \cite{murali2021same} method and reaches near-perfect task success rate and task recall in three testing tasks. Furthermore, the ablation study on placing constraints also indicates that the performance of our framework is bounded by how precise the definition of placing constraints will be. Based on our results, we observe that the placing set is the key to the task success. In the future, our goal is to propose an algorithm that can determine the general placing set more precisely for different tasks.

\addtolength{\textheight}{-12cm}   


\bibliography{IEEEexample}

\begin{thebibliography}{10}

\bibitem{depierre2018jacquard}
A.~Depierre, E.~Dellandr{\'e}a, and L.~Chen, ``Jacquard: A large scale dataset
  for robotic grasp detection,'' in {\em 2018 IEEE/RSJ International Conference
  on Intelligent Robots and Systems (IROS)}, pp.~3511--3516, IEEE, 2018.

\bibitem{tobin2017domain}
J.~Tobin, R.~Fong, A.~Ray, J.~Schneider, W.~Zaremba, and P.~Abbeel, ``Domain
  randomization for transferring deep neural networks from simulation to the
  real world,'' in {\em 2017 IEEE/RSJ international conference on intelligent
  robots and systems (IROS)}, pp.~23--30, IEEE, 2017.

\bibitem{james2019sim}
S.~James, P.~Wohlhart, M.~Kalakrishnan, D.~Kalashnikov, A.~Irpan, J.~Ibarz,
  S.~Levine, R.~Hadsell, and K.~Bousmalis, ``Sim-to-real via sim-to-sim:
  Data-efficient robotic grasping via randomized-to-canonical adaptation
  networks,'' in {\em Proceedings of the IEEE/CVF Conference on Computer Vision
  and Pattern Recognition}, pp.~12627--12637, 2019.

\bibitem{bicchi2000robotic}
A.~Bicchi and V.~Kumar, ``Robotic grasping and contact: A review,'' in {\em
  Proceedings 2000 ICRA. Millennium conference. IEEE international conference
  on robotics and automation. Symposia proceedings (Cat. No. 00CH37065)},
  vol.~1, pp.~348--353, IEEE, 2000.

\bibitem{redmon2015real}
J.~Redmon and A.~Angelova, ``Real-time grasp detection using convolutional
  neural networks,'' in {\em 2015 IEEE international conference on robotics and
  automation (ICRA)}, pp.~1316--1322, IEEE, 2015.

\bibitem{mahler2017dex}
J.~Mahler, J.~Liang, S.~Niyaz, M.~Laskey, R.~Doan, X.~Liu, J.~A. Ojea, and
  K.~Goldberg, ``Dex-net 2.0: Deep learning to plan robust grasps with
  synthetic point clouds and analytic grasp metrics,'' {\em arXiv preprint
  arXiv:1703.09312}, 2017.

\bibitem{morrison2018closing}
D.~Morrison, P.~Corke, and J.~Leitner, ``Closing the loop for robotic grasping:
  A real-time, generative grasp synthesis approach,'' {\em arXiv preprint
  arXiv:1804.05172}, 2018.

\bibitem{mousavian20196}
A.~Mousavian, C.~Eppner, and D.~Fox, ``6-dof graspnet: Variational grasp
  generation for object manipulation,'' in {\em Proceedings of the IEEE/CVF
  International Conference on Computer Vision}, pp.~2901--2910, 2019.

\bibitem{breyer2021volumetric}
M.~Breyer, J.~J. Chung, L.~Ott, R.~Siegwart, and J.~Nieto, ``Volumetric
  grasping network: Real-time 6 dof grasp detection in clutter,'' in {\em
  Conference on Robot Learning}, pp.~1602--1611, PMLR, 2021.

\bibitem{zhu20216dcgpn}
X.~Zhu, L.~Sun, Y.~Fan, and M.~Tomizuka, ``6-dof contrastive grasp proposal
  network,'' in {\em 2021 IEEE International Conference on Robotics and
  Automation (ICRA)}, pp.~6371--6377, 2021.

\bibitem{zhu2022learn}
X.~Zhu, Y.~Zhou, Y.~Fan, L.~Sun, J.~Chen, and M.~Tomizuka, ``Learn to grasp
  with less supervision: A data-efficient maximum likelihood grasp sampling
  loss,'' in {\em 2022 International Conference on Robotics and Automation
  (ICRA)}, pp.~721--727, 2022.

\bibitem{fang2020learning}
K.~Fang, Y.~Zhu, A.~Garg, A.~Kurenkov, V.~Mehta, L.~Fei-Fei, and S.~Savarese,
  ``Learning task-oriented grasping for tool manipulation from simulated
  self-supervision,'' {\em The International Journal of Robotics Research},
  vol.~39, no.~2-3, pp.~202--216, 2020.

\bibitem{gupta2019relay}
A.~Gupta, V.~Kumar, C.~Lynch, S.~Levine, and K.~Hausman, ``Relay policy
  learning: Solving long-horizon tasks via imitation and reinforcement
  learning,'' {\em arXiv preprint arXiv:1910.11956}, 2019.

\bibitem{kalashnikov2018qt}
D.~Kalashnikov, A.~Irpan, P.~Pastor, J.~Ibarz, A.~Herzog, E.~Jang, D.~Quillen,
  E.~Holly, M.~Kalakrishnan, V.~Vanhoucke, {\em et~al.}, ``Qt-opt: Scalable
  deep reinforcement learning for vision-based robotic manipulation,'' {\em
  arXiv preprint arXiv:1806.10293}, 2018.

\bibitem{lee2021ikea}
Y.~Lee, E.~S. Hu, and J.~J. Lim, ``Ikea furniture assembly environment for
  long-horizon complex manipulation tasks,'' in {\em 2021 ieee international
  conference on robotics and automation (icra)}, pp.~6343--6349, IEEE, 2021.

\bibitem{wang2022robot}
J.-W. Wang, C.-L. Li, J.-L. Chen, and J.-J. Lee, ``Robot grasping in dense
  clutter via view-based experience transfer,'' {\em International Journal of
  Intelligent Robotics and Applications}, vol.~6, no.~1, pp.~23--37, 2022.

\bibitem{detry2017task}
R.~Detry, J.~Papon, and L.~Matthies, ``Task-oriented grasping with semantic and
  geometric scene understanding,'' in {\em 2017 IEEE/RSJ International
  Conference on Intelligent Robots and Systems (IROS)}, pp.~3266--3273, IEEE,
  2017.

\bibitem{kokic2017affordance}
M.~Kokic, J.~A. Stork, J.~A. Haustein, and D.~Kragic, ``Affordance detection
  for task-specific grasping using deep learning,'' in {\em 2017 IEEE-RAS 17th
  International Conference on Humanoid Robotics (Humanoids)}, pp.~91--98, IEEE,
  2017.

\bibitem{lakani2018exercising}
S.~R. Lakani, A.~J. Rodr{\'\i}guez-S{\'a}nchez, and J.~Piater, ``Exercising
  affordances of objects: A part-based approach,'' {\em IEEE Robotics and
  Automation Letters}, vol.~3, no.~4, pp.~3465--3472, 2018.

\bibitem{do2018affordancenet}
T.-T. Do, A.~Nguyen, and I.~Reid, ``Affordancenet: An end-to-end deep learning
  approach for object affordance detection,'' in {\em 2018 IEEE international
  conference on robotics and automation (ICRA)}, pp.~5882--5889, IEEE, 2018.

\bibitem{murali2021same}
A.~Murali, W.~Liu, K.~Marino, S.~Chernova, and A.~Gupta, ``Same object,
  different grasps: Data and semantic knowledge for task-oriented grasping,''
  in {\em Conference on Robot Learning}, pp.~1540--1557, PMLR, 2021.

\bibitem{wen2022catgrasp}
B.~Wen, W.~Lian, K.~Bekris, and S.~Schaal, ``Catgrasp: Learning category-level
  task-relevant grasping in clutter from simulation,'' in {\em 2022
  International Conference on Robotics and Automation (ICRA)}, pp.~6401--6408,
  IEEE, 2022.

\bibitem{simeonov2022se}
A.~Simeonov, Y.~Du, L.~Yen-Chen, A.~Rodriguez, L.~P. Kaelbling,
  T.~Lozano-Perez, and P.~Agrawal, ``Se (3)-equivariant relational
  rearrangement with neural descriptor fields,'' {\em arXiv preprint
  arXiv:2211.09786}, 2022.

\bibitem{jeng2021gdn}
K.-Y. Jeng, Y.-C. Liu, Z.~Y. Liu, J.-W. Wang, Y.-L. Chang, H.-T. Su, and
  W.~Hsu, ``Gdn: A coarse-to-fine (c2f) representation for end-to-end 6-dof
  grasp detection,'' in {\em Conference on Robot Learning}, pp.~220--231, PMLR,
  2021.

\bibitem{huang2015we}
D.-A. Huang, M.~Ma, W.-C. Ma, and K.~M. Kitani, ``How do we use our hands?
  discovering a diverse set of common grasps,'' in {\em Proceedings of the IEEE
  Conference on Computer Vision and Pattern Recognition}, pp.~666--675, 2015.

\bibitem{shan2020understanding}
D.~Shan, J.~Geng, M.~Shu, and D.~F. Fouhey, ``Understanding human hands in
  contact at internet scale,'' in {\em Proceedings of the IEEE/CVF conference
  on computer vision and pattern recognition}, pp.~9869--9878, 2020.

\bibitem{kokic2020learning}
M.~Kokic, D.~Kragic, and J.~Bohg, ``Learning task-oriented grasping from human
  activity datasets,'' {\em IEEE Robotics and Automation Letters}, vol.~5,
  no.~2, pp.~3352--3359, 2020.

\bibitem{simeonov2022neural}
A.~Simeonov, Y.~Du, A.~Tagliasacchi, J.~B. Tenenbaum, A.~Rodriguez, P.~Agrawal,
  and V.~Sitzmann, ``Neural descriptor fields: Se (3)-equivariant object
  representations for manipulation,'' in {\em 2022 International Conference on
  Robotics and Automation (ICRA)}, pp.~6394--6400, IEEE, 2022.

\bibitem{oleynikova2016signed}
H.~Oleynikova, A.~Millane, Z.~Taylor, E.~Galceran, J.~Nieto, and R.~Siegwart,
  ``Signed distance fields: A natural representation for both mapping and
  planning,'' in {\em RSS 2016 Workshop: Geometry and Beyond-Representations,
  Physics, and Scene Understanding for Robotics}, University of Michigan, 2016.

\bibitem{mo2022o2o}
K.~Mo, Y.~Qin, F.~Xiang, H.~Su, and L.~Guibas, ``O2o-afford: Annotation-free
  large-scale object-object affordance learning,'' in {\em Conference on Robot
  Learning}, pp.~1666--1677, PMLR, 2022.

\end{thebibliography}
\bibliographystyle{ieeetr}

\newpage

\end{document}